\documentclass[conference]{IEEEtran}
\IEEEoverridecommandlockouts
\usepackage{cite}
\usepackage{amsmath,amssymb,amsfonts}
\usepackage{algorithmic}
\usepackage{graphicx}
\usepackage{textcomp}
\usepackage{xcolor}
\usepackage{subfigure}
\def\BibTeX{{\rm B\kern-.05em{\sc i\kern-.025em b}\kern-.08em
    T\kern-.1667em\lower.7ex\hbox{E}\kern-.125emX}}

\usepackage{cite}

\usepackage{xcolor}

\begin{document}
%
\title{Is Flash Attention Stable?}

\author{Alicia Golden$^{1,2}$ \enspace Samuel Hsia$^{1,2}$ \enspace Fei Sun$^{3}$ \enspace Bilge Acun$^{1}$ \enspace Basil Hosmer$^{1}$ \enspace Yejin Lee$^{1}$\\ 
Zachary DeVito$^{1}$ \enspace Jeff Johnson$^{1}$ \enspace Gu-Yeon Wei$^{2}$ \enspace David Brooks$^{2}$ \enspace Carole-Jean Wu$^{1}$ 
\\ \\
$^{1}$FAIR at Meta \quad $^{2}$Harvard University \quad $^{3}$Meta
}

\maketitle

\begin{abstract}
Training large-scale machine learning models poses distinct system challenges, given both the size and complexity of today's workloads. Recently, many organizations training state-of-the-art Generative AI models have reported cases of instability during training, often taking the form of loss spikes. Numeric deviation has emerged as a potential cause of this training instability, although quantifying this is especially challenging given the costly nature of training runs. In this work, we develop a principled approach to understanding the effects of numeric deviation, and construct proxies to put observations into context when downstream effects are difficult to quantify. As a case study, we apply this framework to analyze the widely-adopted Flash Attention optimization. We find that Flash Attention sees roughly an order of magnitude more numeric deviation as compared to Baseline Attention at BF16 when measured during an isolated forward pass. We then use a data-driven analysis based on the Wasserstein Distance to provide upper bounds on how this numeric deviation impacts model weights during training, finding that the numerical deviation present in Flash Attention is 2-5 times less significant than low-precision training. 








\end{abstract}


%
\IEEEpeerreviewmaketitle

\section{Introduction}
As machine learning trends to larger and more complex models, the model training process is becoming ever more compute- and resource-intensive. The advent of Generative AI has further pushed the boundaries in model development, with Large Language Models (LLMs) often training for months at a time, across hundreds or thousands of GPUs. Take, for example, LLaMA2's 70-B parameter model, which required 1,720,320 GPU hours to train~\cite{touvron2023llama2}. With such long training jobs, \textit{training instability} has become increasingly problematic. As reported in works such as Google's PaLM model, training instability often manifests itself in the form of loss spikes occurring up to 20 times throughout training~\cite{chowdhery2022palm}. These loss spikes are costly, as they often cause interrupts in the training process, requiring training to stop and restart.

While previous work has examined possible mitigation responses to increase training stability from an algorithm perspective~\cite{molybog2023theory, liu2023understanding, NEURIPS2022_7b97adea, gilmer2021loss}, the fundamental cause of this instability is still unclear. The stochastic nature of model training, combined with the fact that many such effects only manifest at large scale, present distinct challenges to fully understanding the nature of these instabilities. Furthermore, repeatedly training
such large models to isolate contributing factors is often
unfeasible, given the cost and high demand for data center
compute.

One under-explored potential cause of training instability is \textit{numeric deviation}. Numeric deviation between an optimization and its corresponding baseline can lead to the gradual accumulation of errors, which over the course of training have the potential to culminate in loss spikes that require a resetting of the model state~\cite{chowdhery2022palm}. This is challenging to quantify, as training's stochastic nature suggests some level of numeric deviation might be acceptable, yet determining the threshold for when training becomes unstable proves difficult.

In this work, we develop a principled quantitative approach to understanding numeric deviation in training optimizations. Our approach consists of two phases, including (i) developing a microbenchmark to perturb numeric precision in the given optimization, and (ii) evaluating how numeric deviation translates to changes in model weights through a data-driven analysis based on Wasserstein distance. This ultimately allows us to provide an upper bound on the amount of numeric deviation for a given optimization, and helps to contextualize the improvement within known techniques. We aim to use this principled analysis to evaluate different state-of-the-art optimization techniques and identify whether they are likely to introduce unintended instabilities when used to train large models.

As a case study, we analyze the state-of-the-art optimization technique Flash Attention~\cite{dao2022flashattention}, and quantify the potential numeric deviation introduced. Flash Attention is a widely-adopted technique used to speed up the attention mechanism, often considered a system bottleneck in transformer models~\cite{vaswani2023attention}. However, while offering increased speedup and reduced memory accesses, Flash Attention depends on algorithm optimizations that have the potential to contribute to increased numeric deviation. Specifically, we hypothesize that the addition of rescaling factors could introduce unintentional approximation that leads to numeric tradeoff, which could later impact training stability. We analyze Flash Attention in the context of multi-modal Text-to-Image workloads in order to determine the potential significance of numeric deviation between Flash Attention and its baseline.

Ultimately, we introduce a framework to quantify the numeric deviation of training optimizations and their downstream impacts. Our key contributions are as follows:
\begin{itemize}
    \item \textbf{We design a microbenchmark to isolate the impact of numerical precision on numeric deviation}. Our microbenchmark serves as a technique to measure and quantify numeric deviation resulting from traditionally black-box optimizations such as Flash Attention. By perturbing aspects not typically available through the provided kernel, we initially find Flash Attention sees roughly an order of magnitude more numeric deviation as compared to Baseline Attention at low numerical precision (BF16).

    \item \textbf{We perform a data-driven analysis based on the Wasserstein Distance metric to contextualize this observed numeric deviation and form an upper bound for the impact on downstream model properties.} In our case study, we are able to bound the impact of this observed numerical deviation, and find that Flash Attention introduces roughly $2-5\times$ less model weight deviation as compared to low-precision training.

\end{itemize}

Our investigations underscore the importance of developing a principled approach to not only quantify, but contextualize, the impact of training optimizations on numeric deviation. By constructing proxies to put this numeric deviation in context, we aim to reason about the likelihood of downstream model effects (i.e., training instability) that are traditionally difficult to measure. 

\section{Background}
\label{background}

As a case study for this work, we analyze the state-of-the-art optimization Flash Attention and its potential numeric deviation from Baseline Attention. The Attention operation has been the focus of myraid optimizations as of late. Attention is currently the main system-performance bottleneck of the Transformer architecture, which is a widely adopted technique in modern machine learning algorithms known for its effectiveness in modeling sequence-to-sequence tasks \cite{vaswani2023attention}.

\subsection{Attention as a System-Performance Bottleneck}
The Attention operation essentially derives a weighted sum that represents how much emphasis a model should place on all previous words when generating a new token \cite{vaswani2023attention}. The key computation in Attention involves matrix-multiplications and softmax. Three representations (Query, Key, Value) are first derived from the input vector, each having dimension $N\times d$ , where $N$  is the sequence length and $d$ is the model dimension. The dot product of the Query and Key values then form a $N\times N$  matrix, which scales quadratically in size with sequence length. This so-called similarity matrix subsequently undergoes a softmax operation, before being multiplied with the Value matrix to complete the computation. The full operation is described below:
\begin{center}
    $Attention(Q,K,V)=softmax(Q\dot K^T / \sqrt{d_k}) V$
\end{center}

Given this quadratic scaling with sequence length, myraid techniques have been proposed to accelerate the Attention mechanism, including system-aware methods such as Flash Attention \cite{dao2022flashattention}.

\subsection{Understanding Flash Attention}
\begin{figure}
    \centering
    \includegraphics[width=0.5\linewidth]{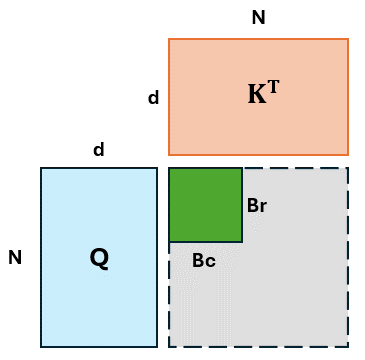}
    \caption{Flash Attention Tiling Operation. Flash Attention uses tiling and recomputation to eliminate the need for the large $N\times N$ similarity matrix. The output of the $QK^T$ dot product is instead calculated by blocks, as shown in green.}
    \label{fig:flash_attn_tiling}
\end{figure}

Flash Attention is a recently proposed technique that is designed to accelerate the Attention bottleneck characteristic of Transformers~\cite{dao2022flashattention}. As an IO aware technique, it aims to minimize the memory overhead of the large $N\times N$ similarity matrix typically used in the attention mechanism. It essentially uses the traditional techniques of tiling and recomputation, along with an online softmax trick, in order to only calculate the matrix one tile at a time \cite{dao2022flashattention}. As shown in Figure \ref{fig:flash_attn_tiling}, the introduction of tiling eliminates the need for the large similarity matrix to be materialized. The block/tile size is defined by $B_c = \lceil M/4d \rceil$ and $B_r = min(\lceil M/4d \rceil, d)$, where M is the size of the SRAM and d is the model dimension. This ensures the block fits inside SRAM, and thus eliminates the need for the large matrix to be loaded/stored from HBM.

However, since the online softmax technique requires global information about the matrix, Flash Attention must incorporate \textit{re-scaling factors} in order to allow for consistent calculations since global information is no longer available when computing on a single block. While introducing relatively minimal overhead, these additional re-scaling factors do introduce extra computation that is calculated per tile.

Flash Attention comes with timing performance speedup as well as more efficient resource utilization; we find that it yields a 14\% speedup of forward + backward pass for our example text-to-image model. However, it has also been hypothesized that the additional computation introduced by rescaling factors of Flash Attention could introduce numeric deviation when used in text-to-image model training, and we aim to investigate this below.

\section{Experimental Methodology}

\begin{figure*}[t!]
    \centering
    \includegraphics[width=0.75\linewidth]{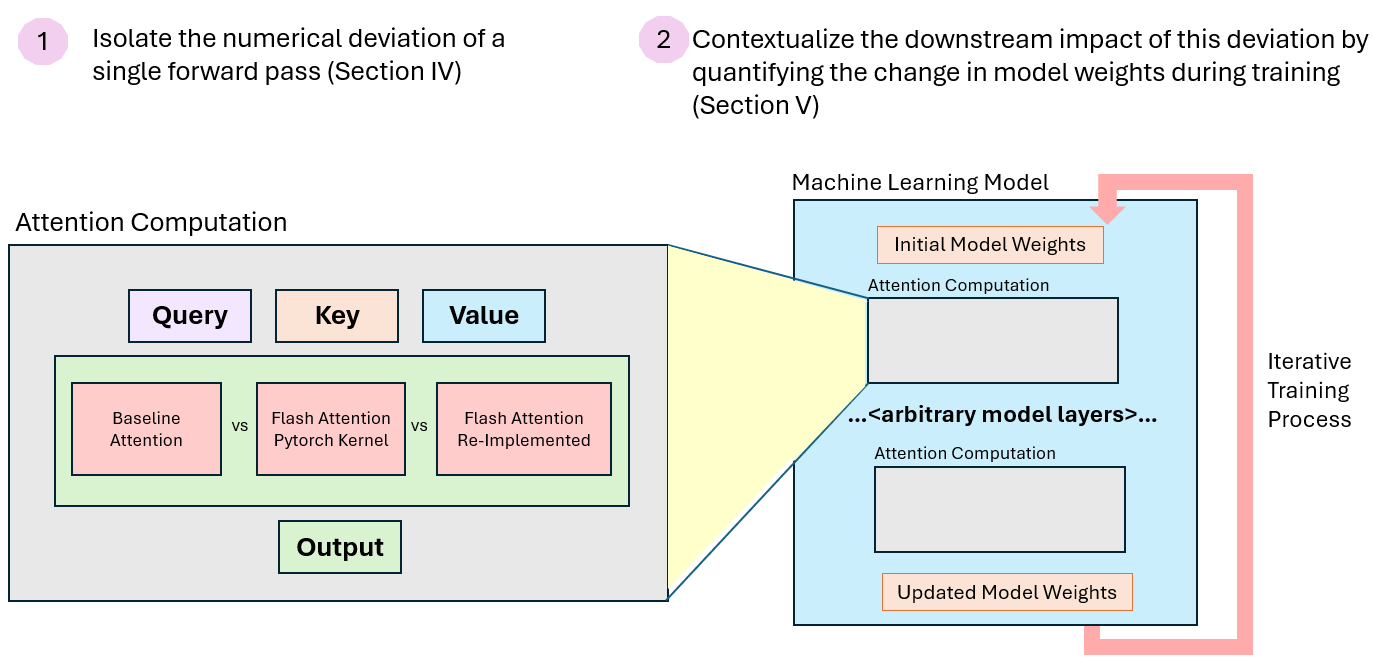}
    \caption{Experimental Methodology. (1) We implement a numerical microbenchmark of the Flash Attention operation, which allows for the experimentation of different numerical precisions, as well as the testing of various optimizations throughout the algorithm. Our framework allows for the direct comparison of the Attention Matrix output between Baseline Attention, Flash Attention, and our numeric re-implementation. (2) We utilize a data-driven procedure to contextualize this numeric difference via examining model weight changes over the course of training.}
    \label{fig:microbenchmark_design}
\end{figure*}

We first develop a microbenchmark to isolate and study the numeric deviation caused by Flash Attention. 
A summary of our microbenchmark design can be found in Figure~\ref{fig:microbenchmark_design}. As shown, we numerically re-implement Flash Attention in order to analyze different numerical precisions and apply potential optimizations at each step of the algorithm, which is not easily done with the original CUDA implementation. This is necessary, as the Flash Attention kernel currently only supports FP16 and BF16 number formats. The Flash Attention kernel is also a wrapped API call of CUDA code, making it challenging to perturb the algorithm to examine the impact on numeric deviation. In contrast, our microbenchmark design allows for varying precision inputs and modifications inside the algorithm. We validate our microbenchmark against the original Flash Attention kernel.

We further devise a technique to compare the output of the Attention matrix at each step during model execution. We modify the model code to compute both Baseline Attention and Flash Attention each time Attention is called, which allows for identical input matrices and an exact output matrix comparison. To contextualize this, we additionally quantify the difference in \textit{model weights} throughout training via identical and independent training runs, using the $max\_difference$ and $Wasserstein\ Distance$ metrics, which are further detailed in Section \ref{weight_differences}. 

For training experiments we use a type of Generative AI workload that converts text inputs to images (i.e., Text-to-Image model). We retrain the model using the Shutterstock dataset and run our experiments across a cluster of NVIDIA 80GB A100 GPUs.

\section{Quantifying Numeric Deviation Through Microbenchmark} \label{forward_pass}

We first analyze the impact of Flash Attention during the forward pass. We utilize our microbenchmark to examine how different numerical precisions impact the output matrix of the Attention calculation, given the same randomly initialized query, key, value vectors.

\subsection{Sweep Numerical Precision} \label{numerical_precision}
\begin{figure}
    \centering
    \includegraphics[width=0.75\linewidth]{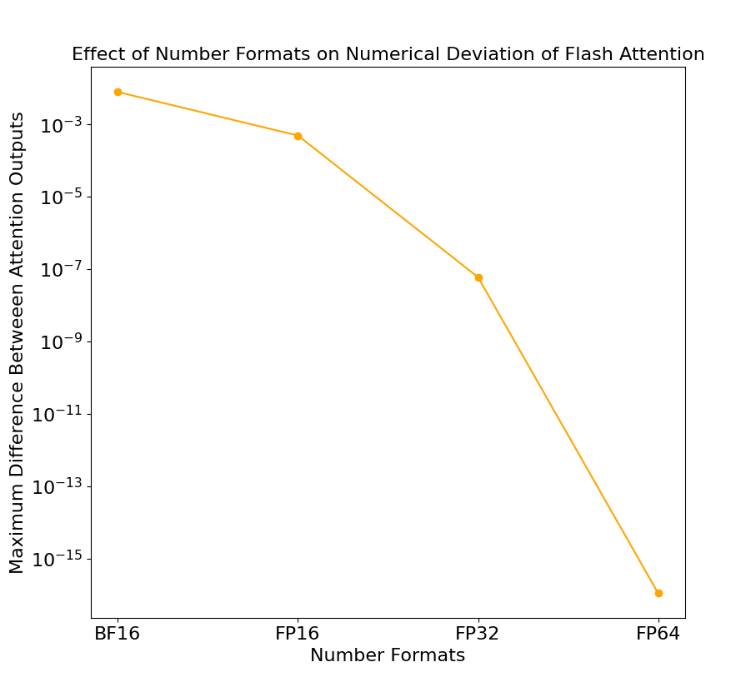}
    \caption{Sweep of numeric precision reveals that there exists a numerical difference between Flash Attention and Baseline Attention, and this varies with numerical precision. As the number format changes from BF16 to FP64, the numeric deviation between Flash Attention and Baseline Attention decreases.}
    \label{fig:number_formats}
\end{figure}

We find that numerical precision has an impact on the output of Flash Attention, causing it to deviate from the output of Baseline Attention. We quantify this through a measure of the maximum difference between attention output matrices taken elementwise, which serves as an upper bound on the possible deviation. As Figure~\ref{fig:number_formats} shows, when using different number formats varying from BF16 to FP64, the numeric deviation between Flash Attention and Baseline Attention decreases with increasing number of mantissa bits. This suggests the numeric difference is a result of approximation inherent with fewer mantissa bits.

\begin{figure}[t!]
    \centering
    \includegraphics[width=\linewidth]{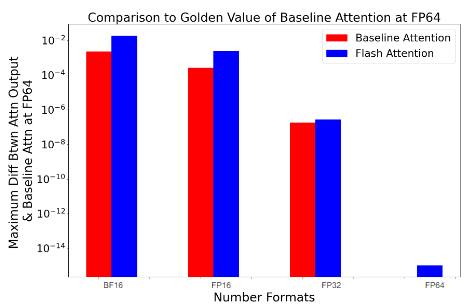}
    \caption{Comparison of Flash Attention at different number formats to Golden Value of Baseline Attention at FP64. We find that Flash Attention sees roughly 10x more numeric deviation as compared to Baseline Attention at BF16.}
    \label{fig:number_formats_alt}
\end{figure}

We then subsequently compare this to the behavior of Baseline Attention. For a common standard of comparison, we set a "golden value" of Attention to be Baseline Attention at FP64. We then compare the maximum difference of the Attention output at different number formats to this golden value, as shown in Figure~\ref{fig:number_formats_alt}. Note we plot the maximum difference between Flash Attention outputs and this golden value (blue bars), while comparing Baseline Attention outputs to this golden value for comparison as well (red bars). We find that Flash Attention sees roughly 10$\times$ more numeric deviation as compared to Baseline Attention at BF16. A detailed discussion on whether this level of deviation is significant can be found in Section \ref{weight_differences}.

\begin{figure}
    \centering
    \subfigure[]{\includegraphics[width=0.49\linewidth]{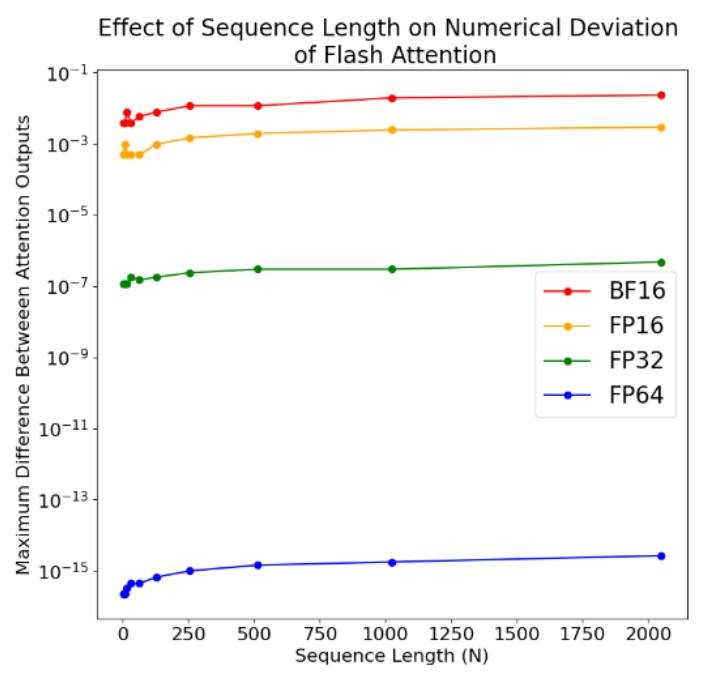}}
    \subfigure[]{\includegraphics[width=0.49\linewidth]{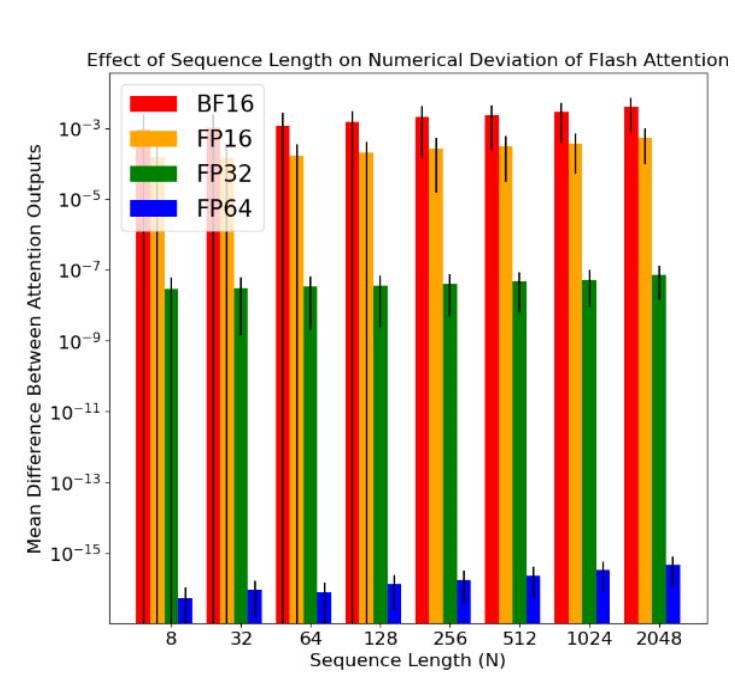}}
        \caption{Impact of Sequence Length on Numerical Deviation of Flash Attention. (a) Increasing sequence length increases maximum difference between Attention matrix outputs (b) Difference between attention output distributions measured with mean and standard deviation reflects a similar trend.}
    \label{fig:sequence_length}
\end{figure}

To analyze this observed numeric deviation further, we sweep the sequence length of our matrices while keeping the tile size and SRAM size the same. As shown in Figure~\ref{fig:sequence_length}, as sequence length increases, the numerical deviation between Flash Attention and Baseline Attention increases, when measured by both (a) the maximum difference upper bound, and (b) the mean and standard deviation of that difference. Since a larger sequence length implies a larger $N\times N$ intermediate matrix that must be tiled while the tile size stays the same, more rescaling is needed. This presents more opportunities for precision errors to accumulate, and thus more deviation.

\subsection{Sweep Algorithm Changes}

\begin{figure*}[t!]
    \centering
    \subfigure[]{\includegraphics[width=0.3\linewidth]{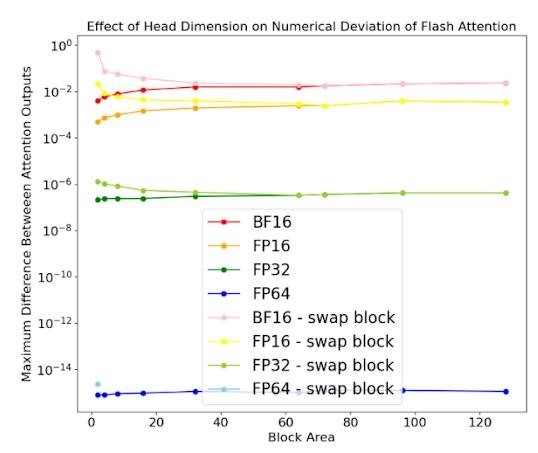}}
    \subfigure[]{\includegraphics[width=0.29\linewidth]{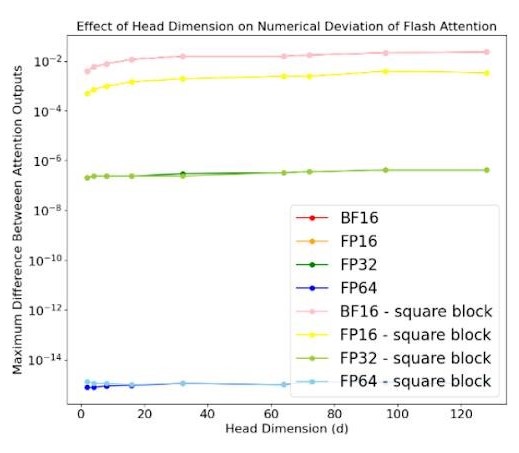}}
    \subfigure[]{\includegraphics[width=0.31\linewidth]{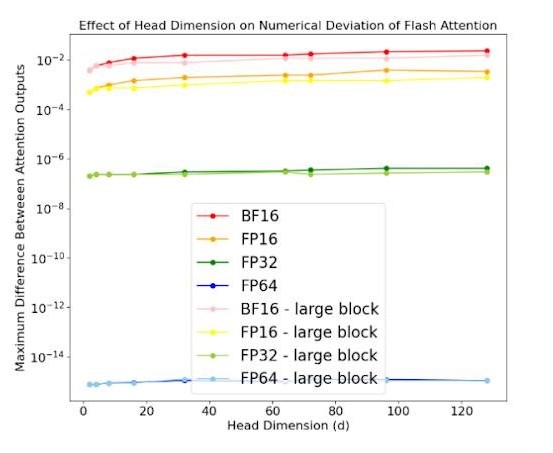}}
    \caption{Algorithm changes and their impact on observed numeric deviation. (a) Swapping the order of block dimensions leads to larger numeric deviation (b) Constraining block sizes to be square does not have a significant impact on numeric deviation (c) Large block sizes lead to smaller numeric deviation. }
    \label{fig:algorithm_changes}
\end{figure*}

We further leverage the microbenchmark design to experiment with different optimizations so we can better understand the effects of this numeric deviation. Figure~\ref{fig:algorithm_changes} shows several algorithm changes and their corresponding impact on observed numeric deviation. For each experiment, we sweep block area (defined according to Bc and Br dimensions introduced in Section \ref{background}), and plot the corresponding maximum difference between Attention matrix outputs. We highlight how this changes at the four precisions analyzed in Section \ref{numerical_precision}. Figure~\ref{fig:algorithm_changes}a shows how swapping the order of the block dimensions leads to larger numerical difference between Flash Attention and Baseline Attention.

Notably, we additionally find that larger block/tile sizes lead to smaller numeric deviation (Figure \ref{fig:algorithm_changes}c). This is because less re-scaling calculations are needed with larger tile sizes, since there are fewer tiles needed to cover the original $N\times N$ matrix. Note that other perturbations such as constraining tile sizes to be square does not have an impact on numeric deviation, since this does not drastically change the number of re-scaling calculations needed to be performed (Figure \ref{fig:algorithm_changes}b).

\section{Contextualizing Numeric Deviation Via Weight Differences} \label{weight_differences}

While Flash Attention may cause numeric deviation of the Attention output during a forward pass, our ultimate goal is to determine whether this has any impact during model training, in order to investigate whether it contributes to training instability. We thus aim to quantify whether Flash Attention changes the model during training --- i.e., if the observed Attention output difference from Section \ref{forward_pass} is reflected in \textit{model weights} that are updated during training.

We utilize two metrics to measure the model weight difference between a model trained with Baseline Attention as compared to Flash Attention. We first calculate $max\_difference$, by finding the absolute value of the difference between weight matrices and taking the maximum to give an upper bound on the deviation, as shown below:
\begin{center}
    $torch.max(torch.abs(flash\_attn - baseline\_attn))$
\end{center}
While $max\_difference$ provides an upper bound on numeric deviation, it fails to take into account the \textit{distribution} of each matrix. We therefore additionally quantify weight differences through Wasserstein Distance, which is a commonly used metric to measure the similarity between tensors~\cite{Panaretos_2019, harma2023accuracy}. While slightly more computationally complex, the Wasserstein metric incorporates information about the shape of the tensor distributions in order measure similarity. The formulation for Wasserstein Distance is outlined below:
\begin{center}
    $W(P,Q) = inf_{g \in \Pi (P,Q)} E_{(x,y) \sim g}[||x-y||]$
\end{center}
Note that lower values indicate higher similarity between matrices.

\begin{figure}[t!]
    \centering
    \includegraphics[width=\linewidth]{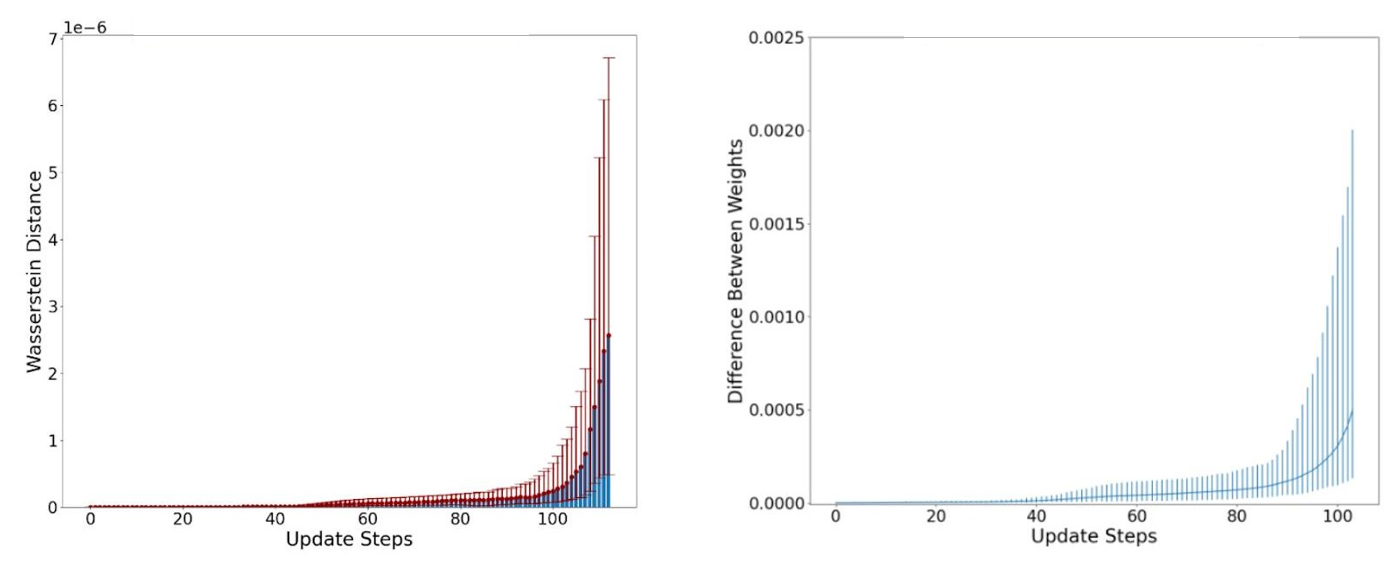}
    \caption{Quantifying how weights of a model with Flash Attention change throughout training as compared to an otherwise identical model using Baseline Attention. Both the metrics of Max Difference and Wasserstein Distance show a similar trend that the difference in model weights increases over the course of training, signifying the models diverge.}
    \label{fig:changing_weights}
\end{figure}

Using these two metrics, we subsequently quantify how model weights change throughout the course of training when Flash Attention is implemented as compared to Baseline Attention. As shown in Figure \ref{fig:changing_weights}, the incorporation of Flash Attention does in fact change the model weights throughout training, as measured by both Wasserstein Distance and Max Difference, and as training continues, this difference only increases. This suggests that models trained with Flash Attention converge to a different model than an identical one trained with Baseline Attention.

However, training is a stochastic process and some model architecture changes can yield comparable results in terms of downstream effects and accuracy. Thus, even if the model weights differ between a model trained with Flash Attention and Baseline Attention, is this significant? Training models to completion and evaluating accuracy is a costly and resource-intensive task, especially for large models where training takes on order of months. We therefore formulate a proxy to understand (a) \textit{how significant these weight changes are?} and (b) \textit{can we put this into context with standard weight changes in other widely-adopted training optimizations?}

To achieve this, we devise a series of experiments to compare how weight differences change over the course of training under different scenarios. In addition to comparing training runs with Flash and Baseline Attention, we quantify the weight difference of identical training runs where weights are initialized to different random values at the beginning of training. This provides a bound, since random weight initialization is a commonly-used technique \cite{scabini2022improving}, and typically produces equivalent results. Furthermore, we also measure the change in model weights of models trained with different precisions. Numeric precision (i.e., FP16 vs FP32) has potential to cause changes downstream, and this serves as an upper bound for determining the significance of Flash Attention weights.

\begin{figure}[t!]
    \centering
    \includegraphics[width=\linewidth]{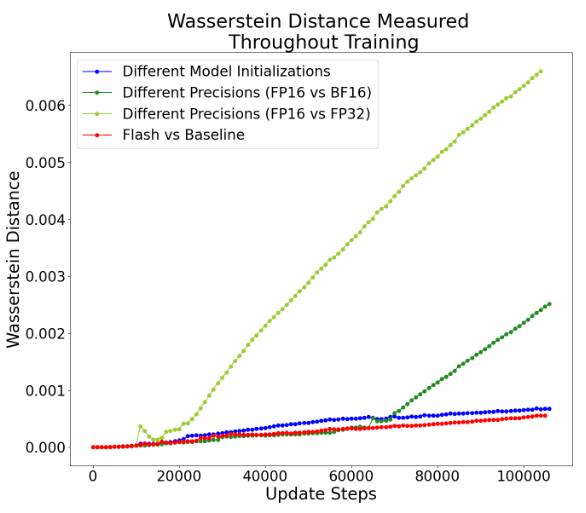}
    \caption{Relative weight differences over training measured using Wasserstein Distance metric. Training a model with Flash Attention causes 2-5x less model weight deviation than low-precision training.}
    \label{fig:relative_weights}
\end{figure}

Figure~\ref{fig:relative_weights} shows the relative weight differences over the course of training as measured using Wasserstein Distance. We find that the rate of change of weight deviation for a model using Flash Attention is comparable or less than the deviation from a different model initialization (note the slope of red and blue curves). Furthermore, we see that the rate of change of weights when using FP16 vs FP32 is higher and more variable than the rate of change for different model initializations. These results provide a proxy to suggest that although numeric deviation occurs with Flash Attention, it is bounded by random model initialization and low-precision training, and introduces roughly $2-5\times$ less model weight deviation as compared to low-precision training.

\section{Discussion and Future Work}

Our work takes the first step towards addressing the question, \textit{Is Flash Attention Stable}, yet there is still more work that needs to be done before drawing a conclusive answer.  Since training instability is challenging and costly to isolate, we explore numeric deviation, which is hypothesized to be a potential cause. Through our principled numeric deviation analysis, we make progress towards this goal by developing a framework to quantify numeric deviation and develop proxies to bound the impact of this deviation in terms of model weights. However, ultimately linking this numeric deviation back to training instability requires further investigation. 

Our numeric quantification methodology opens up a broader set of inquiries, including understanding how various other optimizations impact numeric deviation. Although this analysis is focused on Flash Attention, future work should aim to widen the scope, and investigate additional training optimizations and their corresponding numeric deviation from the appropriate baseline. For example, investigating the numeric deviation caused by the Winograd Algorithm compared to traditional convolution, or various other Attention optimizations, kernel fusion techniques, etc.

More broadly, this work raises a larger set of research questions that deserve attention, specifically regarding training instability. It would be interesting to investigate, for example, the exact point in training where loss spikes occur, and then relate this to the measured weight deviation at this same point. In general, further developing proxies to understand training instability is critical, given the costly nature of these experiments. The following aspects would be interesting to explore as well:

\textit{Training Instability and Reliability.} Training instability leads to interruptions in training, often in the form of loss spikes. However, this is not the only cause of interruptions. Hardware failures, for example, also contribute to the starting/stopping of model training across datacenters. Future work should investigate the relationship between this hardware reliability, checkpointing, and instability.

\textit{Training Instability and System Overhead.} We are additionally interested in quantifying the system overhead that comes with queuing/requeuing training jobs after a loss spike has occurred. The additional overheads that come with these instabilities only magnify when training at-scale, and thus are important to quantify. 

\textit{Training Instability and Sustainability.} Frequent interruptions to the training procedure cause major power swings across the datacenter as well. For example, interruptions can cause significant surges in power consumption when training resumes. This has sustainability implications for designing low-carbon, power-efficient infrastructure for datacenters.

\section{Conclusion}
In this work, we develop a principled approach to understanding the effects of numeric deviation, and develop proxies to put observations into context when downstream effects are otherwise challenging to measure. We examine Flash Attention as a case study, and make progress towards quantifying numeric deviation. We hope that sharing our methodology will allow others to investigate future research questions in a similar manner, and encourage others to investigate this challenging problem of training instability.

\ifCLASSOPTIONcaptionsoff
  \newpage
\fi



%

\bibliographystyle{IEEEtranS}
\bibliography{main}

%




\end{document}